\PassOptionsToPackage{bookmarks=false}{hyperref}
\documentclass[conference]{IEEEtran}

\usepackage[a-2u]{pdfx}
\usepackage[utf8]{inputenc}
\usepackage{lmodern}
\usepackage[T1]{fontenc}
\usepackage{xcolor}
\usepackage{graphicx}

\usepackage{microtype}
\usepackage{listings}
\definecolor{commentcolor}{rgb}{0.5,0.5,0.5}
\definecolor{darkgreen}{rgb}{0.09, 0.45, 0.27}
\definecolor{bgcolor}{rgb}{0.99,0.99,0.99}
\lstdefinestyle{rules}{
    morekeywords={if,else,not,rule,condition,action,allow,pred,isAboveThreshold,isBelowThreshold,hasRightValue1D,hasRightValue2D,hasRightCategories},
    breakatwhitespace=true,           % sets if automatic breaks should only happen at whitespace
    showstringspaces=false,
    columns=fullflexible,
    tabsize=2,%
    breaklines=true,
    captionpos=b,
    numbers=left,
    numbersep=5pt,
    numberstyle=\sffamily\scriptsize\lsstyle,
    xleftmargin=0.05\columnwidth,
    commentstyle=\color{commentcolor},
    keywordstyle=\color{blue},
    stringstyle=\color{darkgreen},
    moredelim=**[is][\color{blue}]{@}{@},
    moredelim=**[is][\sl]{!}{!},
}
\newcommand{\longlstinline}[1]{{\small\textsf{#1}}}
\usepackage{subcaption}
\usepackage{footmisc}
\usepackage{paralist}
\usepackage{tabularx,colortbl}
\usepackage{array}

\usepackage{amsmath}
\renewcommand{\theequation}{R\arabic{equation}}

\usepackage{chngcntr}

\begin{document}
% The following line is necessary to remove dashes (------) for repeated authors in the references section
\bstctlcite{IEEEexample:BSTcontrol}
\title{Towards fuzzification of adaptation rules in self-adaptive architectures}

\author{
\IEEEauthorblockN{Tomáš Bureš$^1$, Petr Hnětynka$^1$, Martin Kruliš$^1$, Danylo Khalyeyev$^1$\\Sebastian Hahner$^2$, Stephan Seifermann$^2$, Maximilian Walter$^2$, Robert Heinrich$^2$}
\IEEEauthorblockA{$^1$\textit{Charles University, Faculty of Mathematics and Physics, Prague, Czech Republic} \\
Email: \{bures, hnetynka, krulis, khalyeyev\}@d3s.mff.cuni.cz}
\IEEEauthorblockA{$^2$\textit{Karlsruhe Institute of Technology (KIT), Germany} \\
Email: \{sebastian.hahner,stephan.seifermann,maximilian.walter,robert.heinrich\}@kit.edu}
}

\maketitle

\begin{abstract}
In this paper, we focus on exploiting neural networks for the analysis and planning stage in self-adaptive architectures. The studied motivating cases in the paper involve existing (legacy) self-adaptive architectures and their adaptation logic, which has been specified by logical rules. We further assume that there is a need to endow these systems with the ability to learn based on examples of inputs and expected outputs. One simple option to address such a need is to replace the reasoning based on logical rules with a neural network. However, this step brings several problems that often create at least a temporary regress. The reason is the logical rules typically represent a large and tested body of domain knowledge, which may be lost if the logical rules are replaced by a neural network. Further, the black-box nature of generic neural networks obfuscates how the systems work inside and consequently introduces more uncertainty. In this paper, we present a method that makes it possible to endow an existing self-adaptive architectures with the ability to learn using neural networks, while preserving domain knowledge existing in the logical rules. We introduce a continuum between the existing rule-based system and a system based on a generic neural network. We show how to navigate in this continuum and create a neural network architecture that naturally embeds the original logical rules and how to gradually scale the learning potential of the network, thus controlling the uncertainty inherent to all soft computing models. We showcase and evaluate the approach on representative excerpts from two larger real-life use cases.
\end{abstract}


\section{Introduction}
\label{sec:introduction}

The recent advances in neural networks helped their proliferation in various other disciplines, the field of software architectures of self-adaptive systems is no exception~\cite{salehie_self-adaptive_2009}. Most importantly, new approaches are emerging that control how architectures of cooperating agents are formed and reconfigured at runtime~\cite{muccini_machine_2019, bures_forming_2020}. 

These approaches employ neural networks to implement the self-adaptation loop (i.e., the MAPE-K loop) that in turn controls runtime decisions in the architecture (e.g., to which service to route a particular request) and the runtime architectural changes (e.g., which services to deploy/un-deploy or reconfigure). 

In typical cases, the neural network is used for the analysis and planning stages of the MAPE-K loop and it replaces the more traditional means of analyzing the situation and deciding on actions, which are often in some form of logical rules (e.g., if-then rules or some form of a state machine with guards and actions)~\cite{muccini_machine_2019,gabor_scenario_2020,van_der_donckt_applying_2020}.

Using a neural network for controlling the operation of a self-adaptive system's architecture inherently means training the network for the situations the self-adaptive system is designed to handle. The training typically requires a large number of examples (training data) in form of observed inputs and expected actions. This is significantly different from the logical rules that the self-adaptive systems have been traditionally described by. Due to this huge gap between the two types of models (neural networks vs. logical rules), it is not easy to evolve an existing self-adaptive system based on some form of logical rules into a new system that uses a neural network for decisions. Seemingly, the only choice is to completely replace the analysis and planning stage with a neural network.

The existing logical rules (especially if the system is well-functioning and tuned to its task) represent a significant body of domain knowledge. Thus, when replacing the logical rules with a neural network, this body of domain knowledge is often lost which is a severe regress. 

Even though if the logical rules are used to generate expected actions in the training data for the neural network, it is not easy to train the neural network to provide reliable answers that correspond with the existing rule-based system. The main culprit is that the neural network is often built as a generic black box --- i.e., composed of generic layers (such as a combination of recurrent and dense layers). Thus, the structure of the neural network does in no way reflect the relationships in the domain which the self-adaptive system controls. By this we mean that the neural network is built as generic and its structure does not exploit the existing domain knowledge about the self-adaptive system whose architectural reconfiguration and operation is about to be controlled. 

While this genericity may be advantageous in allowing the neural network to learn completely unanticipated relationships, it may also hinder the ability to learn because it makes the neural networks unnecessarily complex, and potentially increases the uncertainty in the system.

Therefore, replacing a rule-based system with neural networks altogether may in many cases be a too revolutionary change which may degrade the reliability of the system (at least in the short-term perspective) and may raise legitimate concerns since generic neural networks are much less comprehensible and predictable than rules.

In this paper, we assume an existing self-adaptive system described by logical rules. We attempt to answer these research questions: (1) how to endow the existing system with the ability to learn using neural networks, while preserving domain knowledge existing in the logical rules; and (2) how to scale the learning potential, thus allowing the transition from logical rules to the neural network to be gradual.

We address these research questions by introducing a continuum between the existing rule-based self-adaptive system and a self-adaptive system based on a generic neural network. We show how to convert the logical rules to a neural network with a custom structure that reflects 1:1 the original rules and how to gradually add the learning ability to the neural network. 

An important feature of our approach is that it allows for gradual evolution of a self-adaptive system---i.e., it allows one to start with only a moderate level of learning (replicating closely the original rule-based model) and then over time to relax certain areas of decisions (thus allowing for greater potential to learn, but also to make mistakes). This allows one to better control the uncertainty connected with neural networks.

Our approach can be seen as providing a middle ground between naturally understandable logical rules and fully black-box neural network. By encoding the domain knowledge in the structure of the neural network, it preserves relations present in the application domain and provides a kind of gray-box learning solution for self-adaptive software architectures.

\section{Running Examples}
\label{sec:example}
As a particular examples, we utilize two simple yet realistic use-cases. The first one is taken from our former project focused on security in Industry 4.0 settings\footnote{\url{http://trust40.ipd.kit.edu/home/}}.
The second one focuses on a scheduling problem in ReCodEx system\footnote{\url{https://github.com/ReCodEx}}, a real application for evaluation of coding assignments used at our institution.

\subsection{Industry 4.0 Example}

The goal of this use-case is to use a MAPE-K loop to dynamically reconfigure a software architecture of agents (represented by components) at runtime. The architecture defines groups of components that collaborate and operate on a given task. Each group provides access policies that allow the components to perform their tasks. Thus, the access control is intertwined with dynamic changes in the systems architecture. 

The MAPE-K loop dynamically re-establishes the groups of components to deal with situations in the environment --- e.g., when a machine breaks down, the MAPE-K loop establishes a group of components that communicate and collaborate to fix the machine (i.e., it performs runtime architecture reconfiguration). It also gives them necessary access rights to be able to access the machine logs and to physically enter the room where the machine is located.

In this running example, we pick a particular rule from the larger use-case on architectural reconfiguration in the Industry 4.0 settings and we demonstrate what we mean by the gradual transition from logical rules to the neural network (thus providing a gray-box neural network that embodies the domain knowledge).

\begin{figure}[h]
    \centering
    \includegraphics[width=\columnwidth]{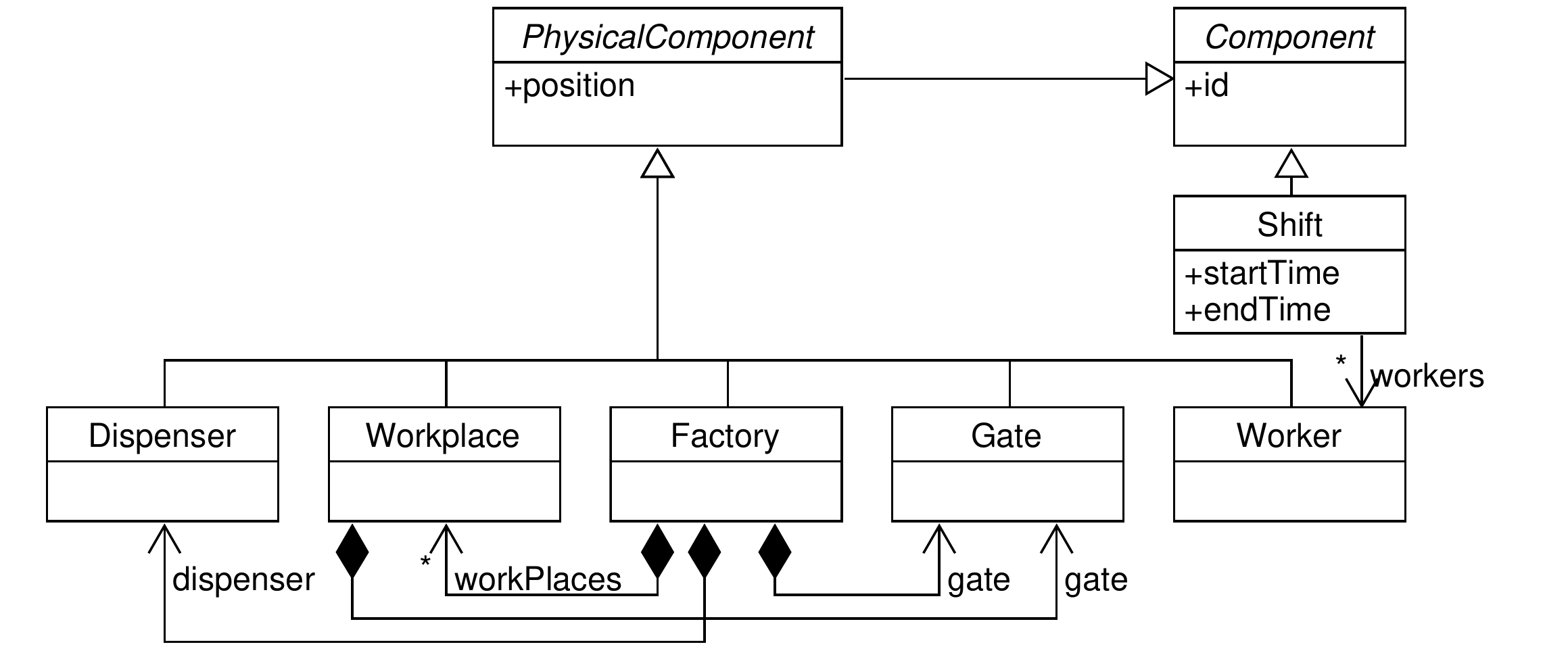}
    \caption{Components of the example}
    \label{fig:domainModel}
\end{figure}

We assume a factory with several workplaces. Work within the factory is organized into shifts, which have their start time and end time, a group of workers, and assigned workplace.
The workers are allowed to enter the factory only at the time close to the particular shift start and they have to leave soon after the end of the shift.
Inside the factory, they have to obtain headgear (i.e., protective equipment for the head) from a dispenser --- without headgear, they are not allowed to enter the workplace.
Also, they are allowed to enter only the assigned workspace and have to enter this workplace close to the shift start time and leave soon after its end.
Figure~\ref{fig:domainModel} summarizes all the components of the example and their relations.

Importantly (as the use-case is taken from the Industry 4.0 domain), the assignment of workers to particular shifts is not static and can frequently change, and also roles of individual workers within the shift can change rapidly, which all leads to changes in the runtime architecture of the system. The access control system in such a factory thus cannot have a fixed configuration of the gates and doors etc. and has to allow for dynamic (situation-based) access control.

In our previous work~\cite{al_ali_toward_2020}, we have created an approach specifying dynamic access control via adaptation rules. We assume the legacy system in the factory to use static access permission (e.g., worker A can access room B). On top of this controlled system, we build the adaptation controller which dynamically establishes the groups of components and modifies the set of static access permissions in the controlled system.

To perform the adaptation, the controller uses adaptation rules in form of condition-action, where the action is adding/revoking and allow permission.
The whole access control is thus treated as self-adaptive architecture, in which an adaptation controller continuously evaluates the conditions
(i.e., access control rules) and executes actions (i.e., assigning/removing individual permissions to elements in the system).

Listing~\ref{lst:access-to-workplace} shows an example of the adaptation rule, which dynamically forms a group components of type worker that are together within a shift and which have the access to the workplace. The group is formed only during the duration of the shift. The example expresses this as a rule which decides whether a particular worker belongs to the group and if so, it gives the worker the access right to the workplace where the shift takes place.

The structure of the adaptation rule has three parts.
First, there can be declared data fields (in this particular example, only a single field initialized to the shift of the particular worker---line~\ref{inlst:shift}).
Then, there is a \textit{condition}, which defines situation when the rule is executed.
The particular rule's condition can be rephrased as follows.
To allow a worker to enter the workplace, the worker needs to be already at the correct workplace entrance (termed workplace gate---line~\ref{inlst:at-gate}), needs to have headgear (line~\ref{inlst:has-headgear}), and needs to be there at the right time (i.e., during the shift or close to its start and end---line~\ref{inlst:during-shift}).
Finally, there is an \textit{action} that declares what has to be executed---in this case assignment of the \textit{allow} rights to the worker for the particular workplace (line~\ref{inlst:assign-allow}).

\begin{lstlisting}[escapechar=|, breaklines=true, style=rules, caption=Access to workplace rule, label=lst:access-to-workplace]
rule AccessToWorkplace(worker) {
    shift = shifts.filter(worker in shift.workers)|\label{inlst:shift}|
    condition {
      duringShift(shift) &&|\label{inlst:during-shift}|
        atWorkplaceGate(worker, shift.workplace) &&|\label{inlst:at-gate}| 
          hasHeadgear(worker)|\label{inlst:has-headgear}|
    }
    action { allow(worker, ENTER, shift.workplace)|\label{inlst:assign-allow}| }
}
\end{lstlisting}

The helper predicates \lstinline{atWorkplaceGate}, \lstinline{hasHeadgear}, and \lstinline{duringShift} are declared in Listing~\ref{lst:in-has}.

The \lstinline{duringShift} tests whether the current time is between 20 minutes (i.e., 1200 seconds) before the start of the shift and 20 minutes after the end of the shift.
The \lstinline{NOW} is a global variable containing the current time.

The \lstinline{atWorkplaceGate} predicate mandates that the position of the worker has to be close (in terms of Euclidean distance) to the gate of the respective workplace at which the worker is supposed to work.

The predicate \lstinline{hasHeadgear} checks whether the worker retrieved the headgear from the dispenser. 
To check this, we assume that each worker is associated with a list of events relating to the worker (the \lstinline{events} data field of the worker---line~\ref{inlst:events} in Listing~\ref{lst:in-has}). 
Retrieving and returning the headgear are events that would be registered in the list of events upon performing the respective actions. 
Thus, the check whether the worker has the headgear is in our example performed by verifying that after filtering only the two events from the list (line~\ref{inlst:filter}), the latest event is \lstinline{TAKE_HGEAR} (we sort the filtered events in the descending order---line~\ref{inlst:sort}---and then the latest one is the first event in the list---line~\ref{inlst:first}).

\begin{lstlisting}[escapechar=/, breaklines=true, style=rules, caption=Helper functions, label=lst:in-has]
pred duringShift(shift) {
	shift.startTime - 1200 < NOW  && shift.endTime + 1200 > NOW
}

pred atWorkplaceGate(worker, workplace) {
    sqrt((workplace.gate.posX - worker.posX) ^ 2 + 
      (workplace.gate.posY - worker.posY) ^ 2) < 10
}

pred hasHeadgear(worker) {
    worker.events/\label{inlst:events}/
     .filter(event -> event.type == 
                   (TAKE_HGEAR || RET_HGER))/\label{inlst:filter}/
     .sortDesc(event -> event.time)/\label{inlst:sort}/
     .first().type == TAKE_HGEAR/\label{inlst:first}/
}
\end{lstlisting}

% ----------------------------------------------------------------------------

\subsection{Coding Assignments Example}

The second example was adopted from the ReCodEx system for evaluation of coding assignments being used at our university which works as follows. The teachers prepare coding assignments which can be evaluated automatically via automated tests. These assignments are given to students who solve them and submit their solutions as source codes. ReCodEx compiles the solutions and executes the test suite prepared by the teacher.

For the purpose of this paper, we have selected a particular problem related to workload distribution over available resources. Since executing the tests is often computationally demanding, the ReCodEx system runs multiple evaluation services (workers) in a private cluster. When a solution is submitted by a student, new evaluation job is created and has to be assigned to one of the workers. The strategy used to distribute jobs obviously affects how long the student has to wait for the test results. Not surprisingly, there is always a spike in the utilization just before the assignment deadline. At this point, the cluster is overutilized and the prioritization on jobs becomes important. The goal becomes to properly prioritize the jobs and route them to the respective workers in the cluster, such that the students do not complain too much about the ReCodEx system being slow.

In line with the method presented in this paper, we model the decisions in the systems using rules. For the sake of simplicity, we use the rules to categorize the jobs to two types: \emph{fast} and \emph{slow}. The fast jobs correspond to assignments that the students expect to get tested fast and thus they actively wait for the results. The slow jobs correspond to assignments, which the students are likely to expect to take some time to get tested (e.g., tens of minutes) so they shift their attention to another task and come back for the results later. 

To get some estimate of how long it could take to test the solution of a particular assignment, we use the reference solution provided by the teacher and measure the time it takes for the solution to get tested and an explicit time limit set for the test of the particular assignment. This gives us the following (strict) rule deciding on whether the job is fast or slow.

\begin{lstlisting}[escapechar=!, breaklines=true, style=rules, caption=Predicate whether a job is deemed slow, label=lst:in-has]
	pred isSlow(job) {
		job.refSolutionDuration > 60 || job.timeLimit > 300
	}
\end{lstlisting}

However, we do not know how long it will take for the student's solution to get tested a priori (as this depends on the quality of the code submitted by the student) and we also do not know where the threshold in the expectations of the students is --- i.e., whether they wait for completion or not. Thus the explicit thresholds ($60$ and $300$) in the rule are simply an educated guess.

\section{Adding Learning Capacity to a Self-Adaptive System}
\label{sec:architecture}

The problem with the logical rules that we presented in the previous section is that these rules are too strict and do not leave any leeway for learning. As already explained in the introduction, a neural network is on the other hand a~structure that can learn, but it is too generic in its usual form. In this section, we outline our approach to find a middle-ground between the strict logical rules and the generic neural network.

In line with our research questions, the approach we present below 
\begin{inparaenum}[(i)]
\item endows an existing self-adaptive system with the ability to learn using neural networks, while preserving domain knowledge captured in the existing logical rules; and 
\item allows one to scale the learning potential and allows for a gradual transition from logical rules to the neural network.
\end{inparaenum}

The main idea of our approach unfolds in three stages:
\begin{enumerate}
\item We systematically apply manual rewriting steps that operate on the level of logical rules and gradually change the strict logical formula to use predicates and relations that allow for learning. Such relaxed logical rules are based on predicates and relations whose semantics relies on trainable parameters (in the same way a neural network has trainable parameters---typically called weights). 
\item We apply an automatic step that generates a custom neural network architecture that reflects the relaxed logical rules and contains the trainable parameters as trainable weights.

\item We use traditional neural network training using stochastic gradient descent to pre-train the trainable parameters. 
\end{enumerate}

The result is a neural network with a custom neural network architecture that corresponds 1:1 to the original structure of the logical formula. This neural network is pre-trained to match outputs of the original logical formula. However, being a neural network, it can be further trained by examples obtained anytime later from runtime.

To have the pre-training data, we assume that there are sample traces of input data to the system (either from historic data, simulator, or just random sampling). We use the original strict logical rules over the input data to provide the ground truth (i.e., expected inputs) that are used in the supervised learning of the neural network.

The key novelty of our approach is that the designer can freely choose how far they want to go with relaxing the original logical rules. As such, it is possible to relax the logical rules only a bit and thus allow only for a small amount of learning (at the benefit of needing only less learning data and having a system that is more predictable), or relax the logical rules significantly and thus allow for a fair amount of learning (at the cost of requiring more data and having a system which is a bit less predictable). In reality, there is a continuous space between the strict logical rules and the generic neural network in which the designer can navigate by choosing which parts of the logical rules to relax and to what extent. For instance, as it will be shown later in the text, the developer can decide to make the time interval defined by the \lstinline{duringShift} predicate trainable, such that it can be automatically adjusted (i.e., made more permissive or more strict) based on when in the typical case the workers arrive and leave the workplace. Further, the developer has the ability to specify the learning capacity of each trainable predicate, which in turn determines how many neurons are used for the implementation of the trainable predicate. This all creates infinitely large space of variants between the strict logical rules on one end and the generic neural network at the other.

\subsection{Relaxing the Logical Rules}
In this section, we exemplify stage \#1 of our approach (i.e., the manual rewriting of the logical formula) on the Industry 4.0 example presented in Section~\ref{sec:example}. We show two steps of such manual rewriting to demonstrate how a designer may choose a different level to which the strict logical rules are relaxed and thus allow for learning.

We start with the logical rules shown in Listings~\ref{lst:access-to-workplace} and~\ref{lst:in-has}. As the first level of relaxation, we assume that the designer would like to relax the predicate \lstinline{duringShift} to make it possible to learn the permitted time interval in which the access is allowed. (In the strict formula, the interval is fixed to be 20 minutes before the shift till 20 minutes after the shift.) 

We redefine the \lstinline{duringShift} predicate as shown in Listing~\ref{lst:relaxed-1}. The comparison of \lstinline{NOW} with a particular threshold has been replaced by predicates \lstinline{isAboveThreshold} and \lstinline{isBelowThreshold} respectively. Each of these predicates represents a comparison against a trainable threshold.

The predicates \lstinline{isAboveThreshold} and \lstinline{isBelowThreshold} have three parameters:
\begin{inparaenum}[(1)]
\item the value to test against the learned threshold, 
\item minimum value for the threshold, 
\item maximum value for the threshold. 
\end{inparaenum}

In our example, we assume this threshold should be the same regardless of the shift, thus we use the times relative to the start and the end of the shift.
Particularly, a worker can arrive as soon as one hour before the shift start, i.e., \lstinline{+3600} seconds in line~\ref{inlst:rel1-start}, and thus relative time \lstinline{0} corresponds to one hour before the shift start (as computed by \lstinline{NOW - shift.end + 3600}).
Similarly, a worker can leave as late as one hour after the shift (\lstinline{-3600} seconds in line~\ref{inlst:rel1-end}) and again, one hour after the shift corresponds to relative time \lstinline{0} (as computed by \lstinline{NOW - shift.end - 3600}).
The minimum and maximum values for the threshold correspond to an interval of 10 hours (i.e., 36000 seconds).

\begin{lstlisting}[escapechar=|, breaklines=true, style=rules, caption=Predicate duringShift as learnable interval, label=lst:relaxed-1]
pred duringShift(shift) {
    isAboveThreshold(NOW - shift.start + 3600,|\label{inlst:rel1-start}|
                     min=0, 
                     max=36000) &&
    isBelowThreshold(NOW - shift.end - 3600,|\label{inlst:rel1-end}|
                     min=-36000, 
                     max=0)
}
\end{lstlisting}

The other predicates (i.e., \lstinline{atWorkplaceGate} and \lstinline{hasHeadgear}) stay the same, as does their conjunction in the \lstinline{AccessToWorkplace} rule. 

Note that we combined strict logic predicates with predicates that have the capacity to learn. This shows how only a part of the logical formula can be endowed with the ability to learn while the rest can be kept strict. 
At the same time, we put strict limits on how far the learning can go. In our example, the limits are expressed by the interval of 10 hours which spans from one hour before the shift to one hour after the shift. Regardless what the trainable predicate learns, it cannot exceed these bounds. This is useful if one wants to combine learning with strict assurances to control uncertainty.

As the second level of relaxation, we assume the time of entry, place of entry, and the relation to the last event concerning the headgear is to be learned. Also, contrary to the example in Listing~\ref{lst:relaxed-1}, we assume the time of entry is not just a single interval but can be a number of intervals (e.g., to reflect the fact that workers usually access the gate only at some time before and after the shift).

To capture this, we redefine the predicates \lstinline{duringShift}, \lstinline{atWorkplaceGate}, and \lstinline{hasHeadGear} as shown in Listing~\ref{lst:relaxed-2}. 

The predicate \lstinline{duringShift} is realized using the predicate \lstinline{hasRightValue1D}, which represents a learnable set of intervals. It has four parameters. In addition to the first three which have the same meaning as before (i.e., value to be tested whether it belongs to any of the learned intervals, the minimum and the maximum value for the intervals), there is a fourth parameter \lstinline{capacity} which expresses the learning capacity. The higher it is, the finer intervals the predicate is able to learn. It works relative to the \lstinline{min}/\lstinline{max} parameters, and thus it is unitless. In our approach, the learning capacity determines the number of neurons used for training. The exact meaning of the \lstinline{capacity} parameter is given further in Section~\ref{sec:transformation}.

The predicate \lstinline{atWorkplaceGate} is designed in a similar fashion. However, as the position is a two-dimensional vector, a 2D version of the \lstinline{hasRightValue} predicate is used. The meaning of its argument is the same as in the 1D version used for the \lstinline{duringShift}. A special feature of the \lstinline{atWorkplaceGate} predicate is that it is specific to the workplace assigned to the worker. (There are several workplaces where the work is conducted during the shift. Each worker is assigned to a particular workplace and their access is limited only to that workplace.) Thus, the \lstinline{hasRightValue2D} predicate has to be trained independently for each workplace. This is expressed by the square brackets after the \lstinline{hasRightValue2D} predicate, which signifies that the training of this predicate is qualified by workplace ID. Given the fact that our running example assumes there are three workplaces in a shift, it effectively means that there are three predicates trained, each with data related to it.

The \lstinline{hasHeadGear} predicate is relaxed using the trainable \lstinline{hasRightCategories} predicate. This trainable predicate assumes a fixed-size vector of categorical values (all from the same domain). It learns which combinations of categorical values correspond to true output. It takes three parameters: the vector of categorical values, the number of categories (i.e., size of the domain of the categorical value), and the learning capacity. 

\begin{lstlisting}[breaklines=true, style=rules, caption=2nd relaxation of the predicates, label=lst:relaxed-2]
pred duringShift(shift) {
    hasRightValue1D(NOW - shift.start, min=0, 
                              max=36000, capacity=20)
}
 
pred atWorkplaceGate(worker) {
    hasRightValue2D@[@!worker.workplace.id!@]@(worker.pos, 
                min=(0,0), max=(316.43506,177.88289),
                                        capacity=20)
}
 
pred hasHeadGear(worker) {
    hasRightCategories(
      worker.events.filter(event -> event.type == 
                        (TAKE_HGEAR || RET_HGER))
                   .sortDesc(event -> event.time).take(1), 
      categories=2, 
      capacity=1
    )
}
\end{lstlisting}

\subsection{Construction of Custom Neural Network Architecture}
\label{sec:transformation}
Having shown how to relax the logical formula of a self-adaption rule to give them the capacity to learn, we formalize in this section stage \#2 of our approach (i.e.,	 automatic generation of a custom neural network architecture that reflects the relaxed logical rule).

We transform a logical formula $L(x_1, \dots, x_m)$ to a continuous function $N(x_1, \dots, x_m, w_1, \dots, w_n) \rightarrow \lbrack 0,1\rbrack$ (i.e., the neural network), where $x_1, \dots, x_m$ are the inputs to the logical rule (i.e., current time, position of the worker, and the last event related to the headgear), and $w_1, \dots, w_n$ are trainable weights. Our goal is to construct the function $N$ and train its weights such that $L(x_1, \dots, x_m) \Leftrightarrow N(x_1, \dots, x_m, w_1, \dots, w_n) > 0.5$ for as many inputs $x_1, \dots, x_m$ as possible. We use the symbol $\mathcal{T}$ to denote this transformation from the logical formula $L$ to the continuous function $N$ --- i.e., $N = \mathcal{T}(L)$.

\vspace{2mm}\noindent %
\textbf{Transformation of a non-trainable predicate.}
As we have shown in Listing~\ref{lst:relaxed-1}, our approach allows combination of non-trainable and trainable predicates. A~non-trainable predicate is transformed simply into a function that returns 0 or 1 depending on the result of the original predicate. Formally, we transform a~non-trainable predicate $L_S(x_1, \dots, x_m)$ to function $N_S(x_1, \dots, x_m)$ as follows:

$$
\mathcal{T}(L_S) = 
\begin{cases}
	0       & \quad \text{if not } L_S(x_1, \dots, x_m)\\
	1  & \quad \text{if } L_S(x_1, \dots, x_m)
\end{cases}
$$

\vspace{2mm}\noindent %
\textbf{Logical connectives.}
We transform the conjunction and disjunction as follows. 
$$
\mathcal{T}(L_1 \& \dots \& L_k) = S\left((\mathcal{T}(L_1) + \dots + \mathcal{T}(L_k) - k + 0.5) * p\right)
$$
$$
\mathcal{T}(L_1 \vee \dots \vee L_k) = S\left((\mathcal{T}(L_1) + \dots + \mathcal{T}(L_k) - 0.5) * p\right)
$$
$$
\mathcal{T}(\lnot L) = 1 - \mathcal{T}(L)
$$

\noindent where $S(x)$ is the sigmoid activation function defined as $S(x)=\frac{1}{1+e^{-x}}$, and $p > 1$ is an adjustable strength of the conjunction/disjunction operator. The bigger it is, the stricter the results. Too high values have however the potential to harm the training due to the vanishing gradient problem.

Note also that we deviated from the traditional notion in which conjunction is defined as a product and disjunction is derived using De Morgan's laws. This is because our experiments showed that the conjunctions of multiple operands are close to impossible to train (very likely due to the vanishing gradient problem).

\vspace{2mm}\noindent %
\textbf{Transformation of trainable \emph{is\dots{}Threshold} predicates.}
We transform the predicate $\mathit{isAboveThreshold}(x, min, max)$ to function $N_>(x, w_t)$ and predicate $\mathit{isBelowThreshold}(x, min, max)$ to function $N_<(x, w_t)$ as follows.

$$
\mathcal{T}(\mathit{isAboveThreshold}) = S\left(\left(\frac{x - min}{max - min} - w_t\right) * p\right)
$$
$$
\mathcal{T}(\mathit{isBelowThreshold}) = S\left(\left(w_t - \frac{x - min}{max - min}\right) * p\right)
$$

\noindent where $w_t$ is a trainable weight.

\vspace{2mm}\noindent %
\textbf{Transformation of trainable \emph{hasRightValue\dots} predicates.}
We base these predicates on radial basis function (RBF) networks~\cite{schwenker_three_2001}. We apply one hidden layer of Gaussian functions and then construct a linear combination of their outputs. The weights in the linear combination are trainable. The training capacity (denoted as $c$ in the definitions below) of the predicate determines the number of neurons (i.e., points for which to evaluate the Gaussian function) in the hidden layer. 

We set the means $\mu_i$ of the Gaussian function to a set of points over the area delimited by \emph{min} and \emph{max} parameters of the predicate (e.g., forming a grid or being randomly sampled from a uniform distribution). We choose $\sigma$ parameter of the Gaussian function to be of the scale of the mean distance between neighbor points. The exact choice of $\sigma$ seems not to be very important. Our experiments have shown that has no significant effect and what matters is only its scale, not the exact value. The trainable linear combination after the RBF layer automatically adjusts to the chosen values of $\mu_i$ and $\sigma$.

In the 1D case, we transform the predicate $\mathit{hasRightValue1D}(x, \mathit{min}, \mathit{max}, c)$ to function $N^1_{\simeq}(x, w_{a_1}, \dots, w_{a_c}, w_b)$ as follows:
$$
\mathcal{T}(\mathit{hasRightValue1D}) = S\left(w_b + \sum_{i=1}^{c}w_{a_i} e^{-\frac{(\mu_i - x)^2}{2\sigma^2}}\right)
$$

\noindent where $c$ is the capacity parameter of the predicate, $\mu_i\in\lbrack \mathit{min},\mathit{max}\rbrack$ and $\sigma$ are set as explained above, and $w_{a_1}, \dots, w_{a_c}, w_b$ are trainable weights.

In the 2D case, we transform the predicate $\mathit{hasRightValue2D}(x, \mathit{min}, \mathit{max}, c)$ to function $N^2_{\simeq}(x, w_{a_{1,1}}, \dots, w_{a_{c,c}}, w_b)$ as follows:
$$
\mathcal{T}(\mathit{hasRightValue2D}) = S\left(w_b + \sum_{i=1}^{c}\sum_{j=1}^{c}w_{a_{i,j}} e^{-\frac{|\mu_{i,j} - x|^2}{2\sigma^2}} \right)
$$

\noindent where $\mu_{i,j}\in\lbrack \mathit{min_1},\mathit{max_1}\rbrack \times \lbrack \mathit{min_2},\mathit{max_2}\rbrack$ and $\sigma$ are set as explained above, $x$ is a 2D vector, $|\cdot|$ stands for vector norm, and $w_{a_{1,1}}, \dots, w_{a_{c,c}}, w_b$ are trainable weights.

\vspace{2mm}\noindent %
\textbf{Transformation of trainable \emph{hasRightCategories} predicate.}
We base this predicate on dense neural network with one hidden layer with the number of units equal to the capacity parameter $c$ of the predicate and being activated by ReLU activation function. 

The transformation of the predicate $\mathit{hasRightCategories}(x, m, c)$ to function $N_{\doteq}(x,w^h_{a_{1,1}},\dots,w^h_{a_{c,m}},w^h_{b_1},\dots,w^h_{b_c},w^o_{a_1},\dots,w^o_{a_{c}},w^o_b)$ is defined as follows:

\begin{multline*}
\mathcal{T}(\mathit{hasRightCategories}) = \\ S\left(w^o_b + \sum_{i=1}^{c}w^o_{a_i}\textrm{ReLU}\left(w^h_{b_i} + \sum_{j=1}^{m}w^h_{a_{i,j}}x_j \right)\right)
\end{multline*}

\noindent where $m$ is the size of the input vector $x$, $c$ is the capacity, and $w^h_{i,j}, w^h_b$ are trainable weights of the hidden layer and $w^o_{a_i}, w^o_b$ are trainable weights of the output layer. The ReLU function is defined as $\mathrm{ReLU}(x)=\max(0,x)$.

\subsection{Addressing Classification Problems}\label{sec:classification}

The Industry 4.0 example helped us demonstrate how to relax if-then rules --- i.e., how to transform logical formulas into their relaxed trainable versions. The second example (evaluation of coding assignments) was selected to prove the process of relaxing the rules is not tailored for one scenario only. In case of the second example, we relax the \emph{isSlow} predicate using using the \emph{isAboveThreshold} construct and logical disjunction that we introduced earlier:

\begin{lstlisting}[breaklines=true, style=rules, caption=Predicate isSlow as learnable interval, label=lst:relaxed-isSlow]
pred isSlow(job) {
    isAboveThreshold(job.refSolutionDuration,
                     min=10, max=120) ||
    isAboveThreshold(job.timeLimit,
                     min=10, max=600)
}
\end{lstlisting}

In this case, we operate directly on time periods, so no additional adjustments are needed. The lower limit was set to $10$ seconds as this should be a time all users should be able to wait. Upper bound was set to the double of the original threshold.

The added complexity of this second example lies in the fact the outputs are not exactly defined, but we gather only negative feedback from the users (as they complain when the job gets delayed too much). Although, this might be viewed to a reinforcement learning problem, we have chosen to select another approach that better suits our needs. We have built a simple model that computes possible distributions of jobs among the workers that minimizes negative feedback, thus transformed it into a classification problem where the workers are classes (technical details are explained in the replication package~\cite{git}).

Our objective is to ensure that the fast jobs will not be delayed significantly, so the users will still perceive them as interactive. Towards that end, the evaluation workers were divided into two groups: One group is dedicated for fast jobs, other group prefers slow jobs but can also process fast jobs when idle.

Typical classification neural network has an output layer that contains one neuron for each class and uses \emph{softmax} activation function. To integrate the \emph{isSlow} predicate with a dense network that may learn other aspects of the problem being solved, the output layer needs to be transformed using the following function right before the final \emph{softmax} is applied:

\begin{align*}
o'_s & = o_s + \mathcal{T}(\mathit{isSlow})\\
o'_f & = o_f \cdot (1 - \mathcal{T}(\mathit{isSlow}))
\end{align*}

The $o_s$ stands for output neurons representing slow workers (classes) and $o_f$ represents fast classes. If the \emph{isSlow} predicate evaluates to $1$, it will increase the weight of slow classes and zeroes the weights of fast classes. Otherwise, it will leave the outputs intact.

\subsection{Training the Custom Neural Network}

The neural network function that we derived as the result of the transformations described in Section~\ref{sec:transformation} and Section~\ref{sec:classification} contains trainable weights. We train these weights using supervised learning. We use the traditional stochastic gradient descent optimization. 

The samples for training are taken from existing logs obtained from the system runtime or from a simulation. In the case of the Industry 4.0 example, each sample contains the current time, the worker id, its position and history of events associated with the worker. 
To obtain the true outputs for supervised learning, we exploit the fact that we have the original non-trainable logical formula available. Thus we use it as an oracle for generating the ground truth for training inputs. The exact training procedure is described in Section~\ref{sec:evaluation} where we present our experiments with training the networks. 

After this training step, the neural network function can be used as a drop-in replacement for the logical rules present in self-adaptive system. Being a~neural network, it is able to digest additional samples generated at runtime --- e.g., to learn from situations when the outputs of the system were manually corrected/overridden by human operators.

\section{Evaluation}
\label{sec:evaluation}
We evaluate our approach by comparing the training results of the custom neural network (NN) created by methods proposed in Section~\ref{sec:architecture} with generic NNs comprising one or two dense layers. The complete set of necessary code and data for replicating the evaluation as well as additional experiments that did not fit this paper are available at~\cite{git}.
%publicly available at GitHub~\cite{git}.

\subsection{Methodology and Datasets}
\label{sec:methodology}

For the Industry 4.0 example, we have created two datasets:
\begin{inparaenum}[(a)]
\item \emph{random} sampled dataset, which has been obtained by randomly generating inputs and using the existing logical rules as an oracle; 
\item \emph{combined} dataset which combines data from a simulation and random data. 
\end{inparaenum}

The datasets have been balanced in such a manner that half of the samples correspond to \emph{true} and half to the \emph{false} evaluation of the guard related to access-control adaptation rule \longlstinline{AccessToWorkplace} (described in Section~\ref{sec:example}). Additionally, to obtain more representative results for evaluation, the false cases have been balanced so that every combination of outcomes of the conditions in the top-level conjunction (i.e., \longlstinline{duringShift} and \longlstinline{atWorkplaceGate} and \longlstinline{hasHeadGear}) has the same probability.

The combined dataset combines false cases from the random sampling and true cases from a simulation. The simulation has been performed by our simulator we have developed in the frame of our applied research project Trust4.0\footnote{\url{https://github.com/smartarch/trust4.0-demo}}. 
The reason for the combination of these two sources is to get better coverage for all possible cases when the guard of the adaptation rule evaluates to \emph{false}.

For the code assignments example, we used anonymized job logs from the ReCodEx system. These logs were combined in a cartesian product with all possible states of worker job queues (i.e., how much is each worker busy) and the training outputs were computed using a rule-based algorithm that takes into account also the supposed feedback from the users.

As the baseline generic NNs, we have selected dense neural networks. Given our experiments and consultation with an expert outside our team (a researcher from another department who specializes on practical applications of neural networks), this architecture suits the best the problem in hand. Our setup comprises networks with one or two dense layers of $128$ to $1024$ nodes (in the case of two layers, both layers have the same amount of nodes). The dense layers used ReLU activation and the final layer used sigmoid.

Representing our approach, three versions of NNs were built corresponding to different levels of relaxation: The first two models relaxed only the time condition, one used the \longlstinline{isAboveThreshold} and \longlstinline{isBelowThreshold} approach (as in Listing~\ref{lst:relaxed-1})---denoted as ``time (A\&B)'', the other used \longlstinline{hasRightValue} predicate (similar to Listing~\ref{lst:relaxed-1} but with \longlstinline{hasRightValue} instead of the combination of \longlstinline{isAboveThreshold} and \longlstinline{isBelowThreshold}) -- denoted as ``time (right)''. The last model relaxed all involved inputs (time, place, and headgear events) as outlined in Listing~\ref{lst:relaxed-2} -- denoted as ``all''. For verification purposes we have also built a Tensorflow model with no trainable parameters (i.e., just rewriting the non-trainable logical formulas using the transformation described in Section~\ref{sec:transformation}) that computed the logical rules exactly. Not surprisingly, it achieved $100\%$ accuracy. Being not trainable, we do not report on this further in the text.

All NN models were implemented in Tensorflow\footnote{\url{https://www.tensorflow.org/} (version 2.4)} machine learning framework and trained on our local GPU cluster. Each dataset was divided into training part ($90\%$) and validation part ($10\%$). We have measured accuracy on the validation part only. All experiments were repeated $5\times$ and we are presenting mean values in our evaluation. For the sake of brevity, we do not report in detail on standard deviations since they were very low (less than $0.1\%$ of accuracy in the worst case).

All training sessions used batch size $100$ and $100$ epochs. We have used the Adam optimizer with cosine decay of the learning rate. In the case of the generic NNs, we have used label smoothing to prevent overfitting\footnote{We have been experimenting with Dropout layers as well, but they were causing significant underfitting in our case.}. The custom NN architecture based on our approach did not seem to suffer from overfitting even without label smoothing, thus we did not apply the label smoothing in this case.

\subsection{Industry 4.0 Results}

\begin{figure}[t!]
	\vspace{-12pt}
	\centering
	\includegraphics[width=0.8\columnwidth]{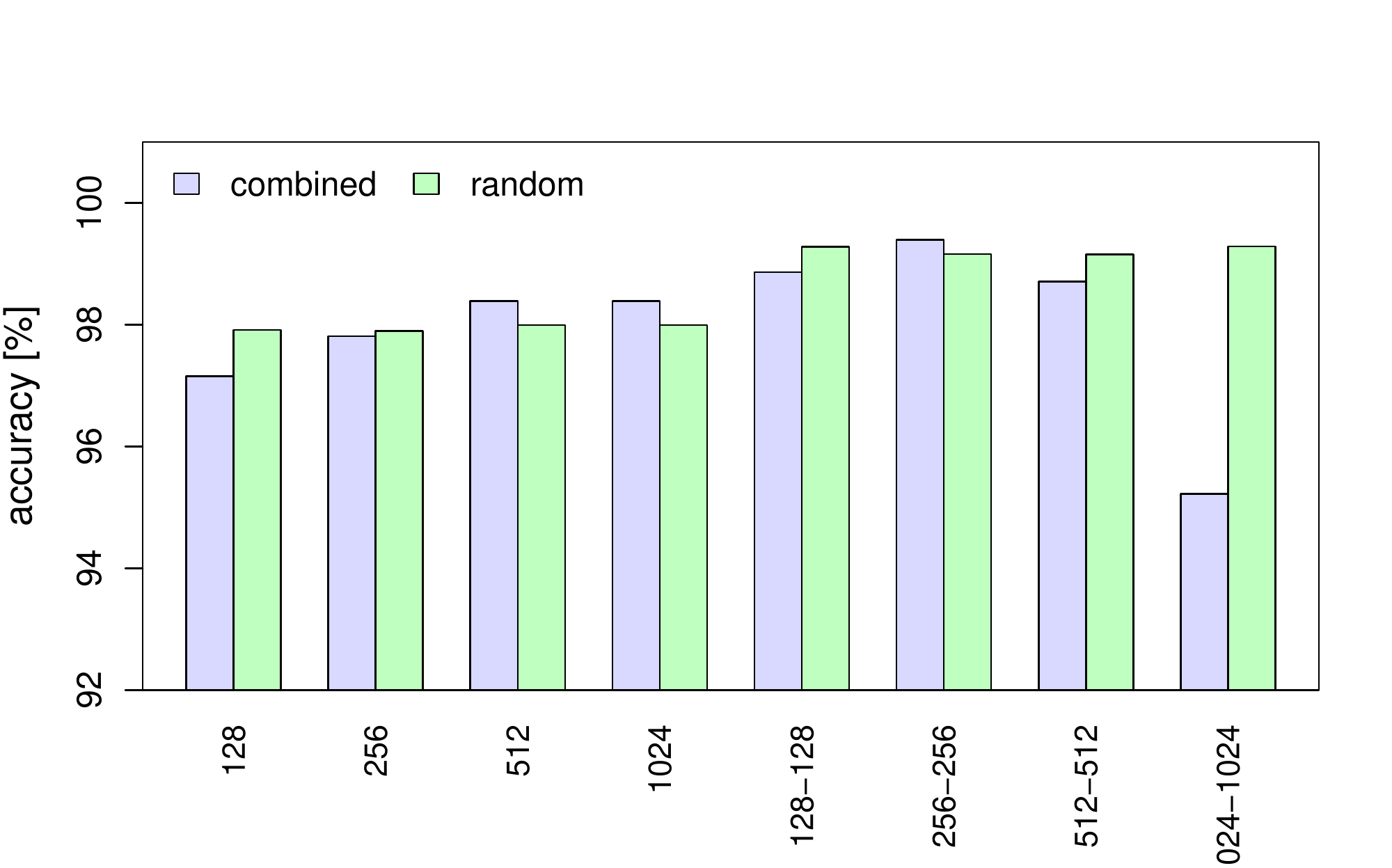}
	\vspace{-5pt}
	\caption{Accuracy of baseline solution (dense networks with one or two layers). The column labels denote widths of dense layers.}
	\label{fig:graph-baseline}
\end{figure}

Figure~\ref{fig:graph-baseline} compares the accuracy of tested configurations after $100$ epochs. The greatest accuracy was observed when two $256$-node layers were used, so we have selected this network as the baseline representative for further comparisons (we refer to this NN simply as the ``baseline''). Smaller NNs suffer from lower capacity, larger nets have stronger inclination to overfitting (especially in the case of combined dataset). It is also worth mentioning that we were experimenting with other sizes and configurations without any improvement.

\begin{table*}[h!]
	\vspace{10pt}
	\setlength{\tabcolsep}{5pt}
	\renewcommand{\arraystretch}{1.5}
	
	\centering
	\begin{tabular}{ l | c c c c }
		& baseline & time (A\&B) & time (right) & all \\ 
		\hline
		Accuracy (random) & $99.159\%$ & $98.878\%$ & $99.999\%$ & $99.978\%$ \\
		Accuracy (combined) & $99.393\%$ & $92.867\%$ & $99.993\%$ & $99.575\%$ \\
		\hline
		Parameters & $68,353$ & $2$ & $21$ & $1,227$
	\end{tabular}
	\vspace{10pt}
	\caption{Comparison of accuracies of individual methods. ``baseline'' is the generic dense NN, the other three correspond to our solution at different levels of relaxation as described in Section~\ref{sec:methodology}.}
	\label{table:accuracies-industry}
\end{table*}

Table~\ref{table:accuracies-industry} presents the measured accuracies on both datasets (random and combined) comparing our approach at different levels of relaxation with the baseline. The last two models outperform the baseline in the terms of accuracy. The \emph{Parameters} line refers to the number of trainable parameters present in each model. Whilst the baseline has many parameters (as it holds two dense layers), our custom NNs have significantly fewer parameters as they are constructed from the domain knowledge of the rules. That makes them less volatile and faster to train.

\begin{figure}[t!]
	\vspace{-12pt}
	\centering
	\includegraphics[width=0.8\columnwidth]{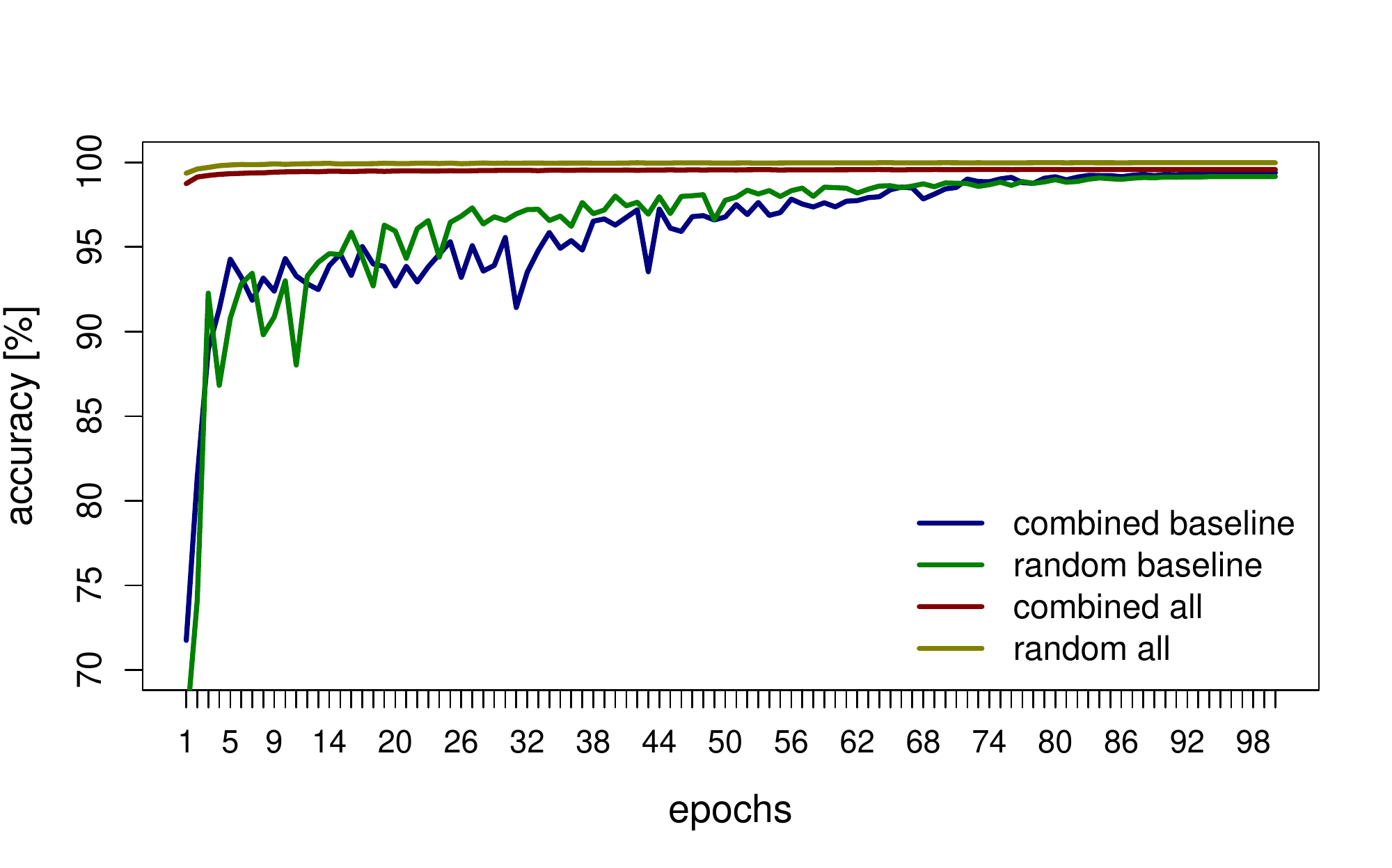}
	\vspace{-8pt}
	\caption{Training speed demonstrated as validation set accuracy measured after every epoch}
	\label{fig:graph-epochs}
\end{figure}

Figure~\ref{fig:graph-epochs} shows the training speed of the models. The generic NNs are much slower in the learning process as they comprise many more training parameters. Our custom NNs reach their peak accuracy within $10$ epochs. In all cases, we can say that $100$ epochs should be sufficient for all models so we have used this as the limit.

\subsection{Coding Assignments Results}

In the coding assignment, we focused mainly on the verification, that adaptation of relaxed rules in classification problem will not affect the training process even with the training using incomplete ground truth (i.e. having only the data about complaints). Since the fuzzification of individual rules have been tested thoroughly on the Industry 4.0 example. The results themselves are very straightforward, the modified neural network model exhibited the same accuracy after training --- about $85\%$. The difference was less than $1\%$ for all tested variations which we consider insignificant.

We should also note that the lower accuracy (with respect to previous examples) is caused by the real-world dataset which contains much more noise than datasets generated from simulations. Furthermore, the multi-class classification problem is more complicated than binary classification. More detailed description of these experiments is provided in the replication package~\cite{git}.

\subsection{Limitations and Threats to Validity}

Though we did our best given the limited scope of the paper, we are aware of several shortcomings of our evaluation. We list the most important ones below. Nevertheless, we believe the evaluation still provides valuable insight into the potential of the approach presented in the paper.

Threats to validity are organized based on the schema in~\cite{runeson_guidelines_2009}, where the validity classes are defined as follows:
\begin{inparaenum}[(i)]
\item construct validity,
\item internal validity,
\item external validity, and
\item reliability.
\end{inparaenum}

\textit{Construct validity:} There is a danger we wrongly constructed our evaluation. As a metric, we measure the accuracy achieved by our approach and compare it with a generic neural network, which we chose as a baseline. This way, we show that our approach does not underperform, even though it uses fewer training weights (thus is more robust) and trains faster. We believe this metric corresponds to the argument we are trying to make in the paper.

\textit{Internal validity:} There is a possible danger that the employed neural network is trained as a black-box. 
Therefore reported improvements gained by our solution might not be results of the proposed construction but rather a result of hidden factor that we are not aware of.
Nevertheless, we are striving to mitigate it by using the exactly same inputs and training procedure when training and evaluating the base-line neural network and the neural network constructed by our approach.

\textit{External validity:} Given the limited scope of the paper, we demonstrated our approach on two adaptation rules only (i.e., \lstinline{AccessToWorkplace} and \lstinline{isSlow}). We selected these rules because they showcase the combination of multiple concerns (time, position, event history, and user preference). However, we are aware of the fact that a larger case study would be needed to identify a more complete set of trainable predicates types (like \lstinline{hasRightValue1D}). The limited complexity of the predicate also does not allow to fully test the limits of what can be learned by a NN. We tried to mitigate this problem to a certain extent by creating additional experiments with synthetically built predicates (e.g., having a conjunction of more than 3 basic conditions). The results of these experiments are part of the results replication package~\cite{git} we made available along with this paper. To ensure the generality of our approach, we also evaluated it on two use-cases from significantly different domains. Especially the second use-case describes an existing system which is used at our school of computer science on daily basis. We also took data for evaluation from the production environment, so the the measurements of time needed to test the students assignments and of the workload generated by the student are real.
To further mature our approach, we plan to evaluate it on further use-cases. This is left for future work.
 
\textit{Reliability:} Another important threat to validity is that we constructed the baseline generic NNs ourselves. To help reduce the bias here, we consulted an expert outside our team to select the best possible architecture fitting the nature of the data. We also performed automated hyperparameter tuning to help us identify the best generic NN architecture, which we eventually used as the baseline.

\section{Related Work}
\label{sec:related}
Using neural networks for the representation of logical formulas and evaluation of fuzzy logical systems is not a new idea.
A kind of calculus for evaluation of propositional logic formulas can be found in~\cite{li_dynamic_2001} together with a proposal for evaluation of fuzzy logical systems.
However, the activation function of neurons in the proposed approach is a simple threshold function. 
More recent approaches (e.g., \cite{riegel_logical_2020,shi_neural_2019}) use the sigmoid function or linear interpolation. We take a similar approach and push this further to practical application aligned with relaxing self-adaptive systems. Also, our approach features a practical approach to logical connectives that are easier to train.

In the domain of adaptive and cyber-physical systems, neural networks and machine learning are used in a number of areas.
Not closely related but rather a big area is the employment of neural networks for anomaly detection---primarily to identify attacks on a system.
The paper~\cite{mohammadi_rouzbahani_anomaly_2020} provides an overview of anomaly detection techniques and machine learning and neural networks cover most of them.
A detailed survey of learning approaches for anomaly detection is in~\cite{luo_deep_2021}. 

There are also a number of closely related approaches that employ neural networks in adaptive systems in their analysis phase of the adaptation cycle.
Typically, these approaches utilize neural networks to predict the best adaptation strategies.
Namely, the approaches are as follows.
In \cite{van_der_donckt_applying_2020}, neural networks are applied during the analysis and planning phase to reduce large adaptation space when the system has multiple adaptation goals and possible optimization strategies.
In our approach, we apply neural networks during the same phases but our goal is to relax strict conditions and thus allow for more flexible adaptation.
In \cite{gabor_scenario_2020}, a whole software engineering framework for adaptive systems is proposed. 
Similar to the previous approach, neural networks are applied during the restriction of the adaptation space to achieve a meaningful system after adaptation.
The approach in \cite{muccini_machine_2019} is slightly different, as neural networks are employed on the boundary of monitoring and analysis phases of the adaptation loop. 
They are used to forecast future values of QoS parameters of a monitored system and thus allow for the progressive selection of the best adaptation strategy.
A similar prediction is used in~\cite{anaya_prediction-driven_2014} to predict values in sensor networks and proactively perform adaptation.
Multiple machine learning algorithms including also neural networks are employed in~\cite{chen_self-adaptive_2017} to create a  dynamic, self-adaptive, and online QoS modeling approach for cloud-based services. 
The approach is again used to predict QoS values and thus allows for optimization of cloud resource utilization.

As described, the approaches above employ neural networks to either reduce the adaptation space or to predict parameters of the system and thus adapt it proactively. 
They differ from our approach as we use neural networks to relax strict conditions in an adaptive system and thus in fact to learn new unforeseen conditions.
Conceptually a similar approach is~\cite{ghahremani_training_2018}, where the authors utilize machine learning approaches to train a model for rule-based adaptation. 
However, instead of neural networks, the approach employs different machine learning approaches like random forest, gradient boosting regression models,
and extreme boosting trees.
Similarly, the paper~\cite{bierzynski_supporting_2019} proposes a proactive learner that is connected to all the phases of the adaptation cycle and continuously updates a learning model.
Nevertheless, the paper discusses mainly the infrastructure and omits details about the used machine learning techniques.
In~\cite{stein_concept_2018}, the authors propose an approach for dynamic learning of knowledge in self-adaptive and self-improving systems, and supervised and reinforcement learning techniques are used.
In~\cite{jamshidi_managing_2016}, machine learning is used to deal with uncertainty in an adaptive system (namely in a cloud controller).
Here, the proposed approach allows users to specify potentially imprecise control rules expressed with help of fuzzy logic (i.e., rules containing specifications like \textit{low} or \textit{high}), and machine learning techniques (namely fuzzy reinforcement learning) are used to learn precise rules.
This approach is in a fact completely opposite to ours where we start with precise rules and via machine learning, we fuzzify them for more flexible adaptation.
A similar approach is in~\cite{zhao_reinforcement_2017}, where reinforcement learning is also employed for the generation of adaptation rules and evolution of them.

\section{Conclusion}
\label{sec:conclusion}

In this paper, we focused on the problem of endowing existing rule-based self-adaptive systems with the ability to learn such that the architectural reconfigurations and runtime decisions in the architecture can be optimized with respect to the data collected by the system. 

We presented an approach that allows one to transform such a system to a custom neural network, whose architecture corresponds 1:1 to the structure of the original logical guards of the adaptation rules. We further showed a couple of examples of trainable predicates that can be used to relax the original strict logical predicate to bring in the ability to learn. 

An important aspect of our approach is that by having the ability to combine non-trainable predicates with trainable ones (and additionally having the ability to set the training capacity of the trainable predicates), one can freely navigate in the continuum between a strict non-trainable system and a relaxed fully trainable system. This makes it possible to perform a gradual transition from the original self-adaptive system to its trainable counterpart while strictly controlling uncertainty connected with introducing machine learning in the system.

The aspect of being able to control the uncertainty connected with the machine learning is the key distinguishing factor of our approach. This stems especially from two facts: (1) The structure of the neural networks produced by our approach has direct relation to the strict logical rules and strict rules (creating an envelope for systems decisions) can be combined with trainable predicates. (2) Our approach yields neural networks that have less neurons than a similarly performing neural network with generic structure (e.g., a multi-layer perceptron network with several hidden dense layers). This makes the networks generated by our approach less prone to overfitting, which may lead to unexpected results in real environment.

In the future work, we aim at extending the base of the trainable predicates to provide a systematic set of predicates for addressing common cases. We are also looking into ways of supporting and the process of gradual rewriting of strict predicates to trainable ones, such as to make this process semi-automatic.

\bibliographystyle{IEEEtran}
\bibliography{paper}

\end{document}